\pdfoutput=1

\documentclass[11pt]{article}

\usepackage[]{acl}

\usepackage{times}
\usepackage{latexsym}
\usepackage{caption}
\usepackage{graphics}
 \usepackage{graphicx}
 \usepackage{adjustbox}
 \graphicspath{ {pics/} }
 \usepackage{multirow}
\usepackage[T1]{fontenc}

\usepackage[utf8]{inputenc}

\usepackage{microtype}

\title{Transformer-based Model for Word Level Language Identification in Code-mixed Kannada-English Texts}

\author{\normalsize Atnafu Lambebo Tonja$^{1}$, Mesay Gemeda Yigezu$^{2}$, Olga Kolesnikova$^{3}$, \\
\textbf{\normalsize Moein Shahiki Tash$^{4}$, Grigori Sidorov$^{5}$, Alexander Gelbukh$^{6}$}\\\\
\normalsize
Instituto Politécnico Nacional (IPN), Centro de Investigación en Computación (CIC), Mexico City, Mexico \\
\footnotesize
\{\texttt{ $^1$alambedot2022, $^2$mgemedak2022, $^5$sidorov, $^6$gelbukh\}@cic.ipn.mx}\\
\footnotesize
\{\texttt{ $^3$kolesolga, $^4$ moein.tash\}@gmail.com}}

\begin{document}
\maketitle
\begin{abstract}
Using code-mixed data in natural language processing (NLP) research currently gets a lot of attention. Language identification of social media code-mixed text has been an interesting problem of study in recent years due to the advancement and influences of social media in communication. This paper presents the Instituto Politécnico Nacional, Centro de Investigación en Computación (CIC) team's system description paper for the CoLI-Kanglish shared task at ICON2022. In this paper, we propose the use of a Transformer based model for word-level language identification in code-mixed Kannada English texts. The proposed model on the CoLI-Kenglish dataset achieves a weighted F1-score of 0.84 and a macro F1-score of 0.61.
\end{abstract}

\section{Introduction}\label{Intro}
In recent years, language identification of social media text has been a fascinating research topic \cite{ansari2021language}. Social media platforms have become more integrated in this digital era and have impacted various people’s perceptions of networking and socializing \cite{tonja2022cic}. This influence allowed different users to communicate via various social media platforms using a mix of texts. NLP technology has advanced rapidly in many applications, including machine translation \cite{tonja2022improving,yigezu2021multilingual, tonja2021parallel}, abusive comment detection \cite{balouchzahi2022mucic}, fake news detection\cite{arif2022cic,truicua2022awakened}, aggressive incident detection \cite{tonja2022detection}, hope speech detection \cite{gowda2022mucic}, and others. However, numerous tools have not yet been created for languages with limited resources or languages with code-mixed data.

Code-mixing is the use of linguistic units—words, phrases, and clauses—at the sentence or word level from various languages. In casual communication, such as social media, it is typically seen. We have access to an enormous amount of code-mixed data because of the various social media platforms that allow individuals to communicate \cite{sutrisno2019beyond}. As a result, automatic language recognition at the word level has become an essential part of analyzing noisy content in social media. It would help with the automated analysis of content generated on social media. Currently, in the area of NLP, different researchers are developing different NLP applications in code-mixed datasets. Some of the applications are code-mixed sentiments analysis \cite{balouchzahi2021cosad}, code-mixed offensive language identification \cite{balouchzahi2021comata}, etc.
We took part in the \textbf{Kanglish shared task} \cite{overviewkanglish}, which aims to identify language at the word level from code-mixed data for Kannada-English texts. For word-level code-mixed language identification tasks, we used Transformer -based \cite{vaswani2017attention} pre-trained language models (PLMs). Our transformer-based model consists of BERT \cite{devlin2018bert} and its three variants. We used PLMs and LSTM models for this word-level language identification task.

This paper discusses a Transformer-based model for word-level language identification in
code-mixed Kannada-English texts for the Kanglish shared task. The paper is organized as follows: Section \ref{Related} describes past work related to this study, section \ref{data} gives an overview of the dataset and its statistics, section \ref{meth} explains the methodology adopted in this study including the algorithms, section \ref{exp} emphasizes on the experimental results and descriptions. Finally, Section \ref{con} concludes the paper.

\section{Related Work}\label{Related}
Currently, solving NLP problems in code-mixed data is getting attention from many researchers. For word-level language identification in code-mixed text, different researchers have suggested various models. \citet{chittaranjan2014word} proposed a Conditional Random Fields (CRF)- based system for word level language identification of code-mixed text for four language pairs, namely, English-Spanish (En-Es), English-Nepali (En-Ne), English-Mandarin (En-Cn), and Standard Arabic-Arabic (Ar-Ar) dialects. The authors explored various token levels and contextual features to build an optimal CRF using the provided training data. The proposed system performed more or less consistently, with accuracy ranging from 80\% to 95\% across four language pairs.

\citet{gundapu2020word} also proposed a CRF based model for word-level language identification in English-Telugu code-mixed data. The authors used feature extraction as the main task for the proposed model. They used POS-tags, length of the word, prefix and suffix of focus word, numeric digit, special symbol, capital letter, and character N-grams (Uni-, Bi-, Trigrams of
words) as features. The proposed CRF-based model had an F1-score of 0.91.

A Support Vector Machines (SVM)-based model for word level language identification of Tamil-English code-mixed text in social media is proposed by \citet{shanmugalingam2018word}. The authors used dictionaries, double consonants, and term frequency to identify features. The proposed SVM model with a linear kernel gave 89.46\% accuracy for the language identification system for Tamil-English code-mixed text at the word level. 

\citet{ansari2021language} proposes transfer learning and fine-tuning BERT models for language identification of Hindi-English code-mixed tweets. The authors used data from Hindi-English-Urdu code-mixed text for language pre-training and Hindi-English code-mixed for subsequent word-level language classification. The authors first pre-trained Hindi-English-Urdu code-mixed text using BERT and fine-tuned the trained model in downstream 
Hindi-English code-mixed word-level language classification. Their proposed model for Hindi-English code-mixed language identification, both pre-training and fine-tuning with code-mixed text, gives the best F1-score of 0.84 as compared to their monolingual counterparts.

\section{Data}\label{data}
\begin{figure}[h!]
    \centering
    \includegraphics{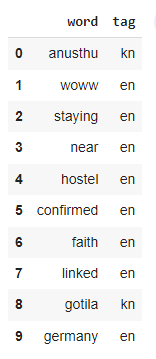}
    \caption{Training samples}
    \label{fig:trainsa}
\end{figure}

During the experimental phase, we used the CoLI-Kenglish dataset \cite{shashirekha2022CoLI} which consists of English and Kannada words in Roman script and are grouped into six major categories, namely, Kannada (kn), English (en), Mixed-language (en-kn), Name, Location and Other. Figure \ref{fig:trainsa} shows some samples from the  dataset used for training.

\subsection{Dataset Statistics}
\begin{figure}[h!]
    \centering
    \includegraphics[width=\linewidth]{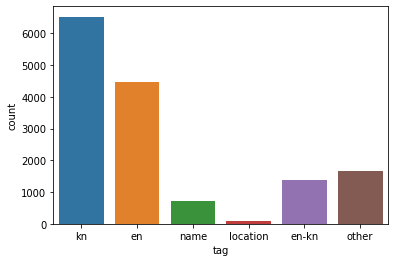}
    \caption{Training data distribution with tags}
    \label{fig:tra}
\end{figure}

\begin{figure}[h!]
    \centering
    \includegraphics[width=\linewidth]{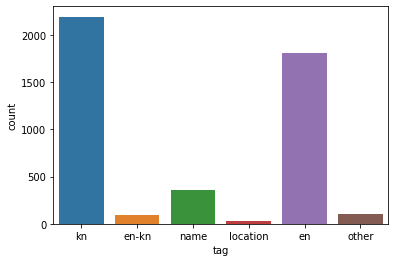}
    \caption{Test data distribution with tags}
    \label{fig:test}
\end{figure}

Figures \ref{fig:tra} and \ref{fig:test} depict the training and test data distribution statistics with their assigned tags. The training dataset is slightly imbalanced: 43.9\% of the words were labeled as \textit{kn}, 30\% were labeled as \textit{en}, 9.28\% were labeled as \textit{en-kn}, 4.76\% were labeled as \textit{name}, 0.68\% were labeled as \textit{location} and 11.2\% were labeled as other. This shows that approximately 73\% of the training dataset was labeled as \textit{kn} and \textit{en}. Similarly, in the test dataset, words tagged as \textit{en} and \textit{kn} take a higher number than the rest of the dataset.

\section{Methodology}\label{meth}
This section presents a description of the data pre-processing, methodology, and models used in this work. We used Transformer based pre-trained language models (PLMs) with the combination of the LSTM model for word level language identification in Kannada-English code-mixed text. We used PLMs in the embedding layer of the LSTM model layer.

\subsection{Pre-processing}
Pre-processing is one of the preliminary steps in training NLP tasks, with the aim of making the training data suitable during the training phase. The dataset provided by the organizers for this task has passed the basic pre-processing steps, and we carried out one pre-processing step to prepare the training data during the experimental phase. We applied label encoding to tags,  to convert the tags into a numeric form. As discussed in section \ref{data}, the dataset contains six tags (\textit{kn, en, en-kn, name, location} and \textit{other}). We converted these tags into numeric values using one-hot encoding.

\subsection{Proposed Experimental Architecture}
\begin{figure}[h!]
    \centering
    \includegraphics[width=\linewidth]{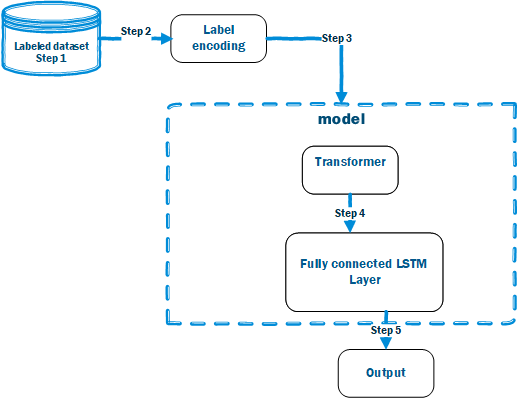}
    \caption{Experimental architecture for Transformer-based model for word level language identification in code-mixed Kannada-English texts}
    \label{fig:arch}
\end{figure}

Figure \ref{fig:arch} shows the experimental architecture of our Transformer-based model for word level language identification in code-mixed Kannada-English texts.
As shown in Figure \ref{fig:arch}, our experimental architecture consists of five steps:
\begin{itemize}
    \item \textbf{Step 1} - preparing labelled data for training, the data set contains \textbf{words} and their \textbf{tags} as discussed in section \ref{data}.
    \item \textbf{Step 2} - we converted the tags into a numeric machine-readable form.
    \item \textbf{Step 3} - after label encoding the representation for each token is fed to transformer layers to obtain contextualized tokens using PLMs. 
    \item \textbf{Step 4} - the embeddings obtained in step-3 are fed into LSTM model to obtain their corresponding language tag.
\end{itemize}
We used the following pre-trained language models (PLMs) in the embedding layer of the LSTM model for our experiment.  \begin{itemize}
    \item BERT \cite{devlin2018bert} - stands for Bidirectional Encoder Representations from Transformers. As the name suggests, it is a way of learning representations of a language that uses a transformer, specifically, the encoder part of the transformer.
    \item mBERT \cite{devlin2018bert} - is a Multilingual BERT, it provides sentence representations for 104 languages, which are useful for many multi-lingual tasks. Previous work probed the cross-linguality of mBERT using zero-shot transfer learning on morphological and syntactic tasks. 
    \item XLM-R \cite{conneau2019unsupervised} - uses self-supervised training techniques to achieve state-of-the-art performance in cross-lingual understanding, a task in which a model is trained in one language and then used with other languages without additional training data.
    \item RoBERTa \cite{liu2019roberta} is a self-supervised transformers model that was trained on a large corpus of English data. This means it was pre-trained on raw texts only, with no human labeling in any way (which is why it can use lots of publicly available data) and an automatic process to generate inputs and labels from those texts.
\end{itemize}

Table \ref{tab:modelpra} shows models used in our experiments and their parameters.
\begin{table*}[h!]

\begin{tabular}{ccccc}
\hline
\multicolumn{1}{l}{\textbf{Model}} & \multicolumn{1}{l}{\textbf{\begin{tabular}[c]{@{}l@{}}Transformer \\ blocks\end{tabular}}} & \multicolumn{1}{l}{\textbf{\begin{tabular}[c]{@{}l@{}}Hidden \\ layer size\end{tabular}}} & \multicolumn{1}{l}{\textbf{\begin{tabular}[c]{@{}l@{}}Self-attention\\  heads\end{tabular}}} & \multicolumn{1}{l}{\textbf{\#Parameters}} \\\hline
bert-base-uncased & 12 & 768 & 12 & 110M \\
bert-base-multilingual-uncased & 12 & 768 & 12 & 110M \\
xlm-roberta-large & 24 & 1024 & 16 & 355M \\
roberta-base & 12 & 768 & 12 & 110M \\\hline
\end{tabular}
\caption{Transformers used in this paper and their parameters }
\label{tab:modelpra}

\end{table*}
\section{Experiments and Results}\label{exp}
This section presents the description of the experimental setups, training parameters, results, and analysis. We conducted four experiments by replacing embedding layers with different pre-trained language models, the results are presented in section \ref{res}.

\subsection{Experiments} \label{expr}
We used Google colab \footnote{https://colab.research.google.com/} for GPU support with the Python programming language. Sci-kit-learn \footnote{https://scikit-learn.org/stable/} and Keras \footnote{https://keras.io/} (with TensorFlow backend) were utilized for the LSTM model, for PLMs we used Hugging Face \footnote{https://huggingface.co/} transformer libraries.
We used PLMs for embedding and the LSTM model as the classifier, To optimize the model, we used an Adam optimizer with a batch size of 64 and a learning rate of 0.0001. We used the maximum number of epochs of 30, with early stopping based on the performance of the validation set. We also used a dropout of 0.2 to regularize the model. 

\begin{figure}[h!]
    \centering
    \includegraphics[width=\linewidth]{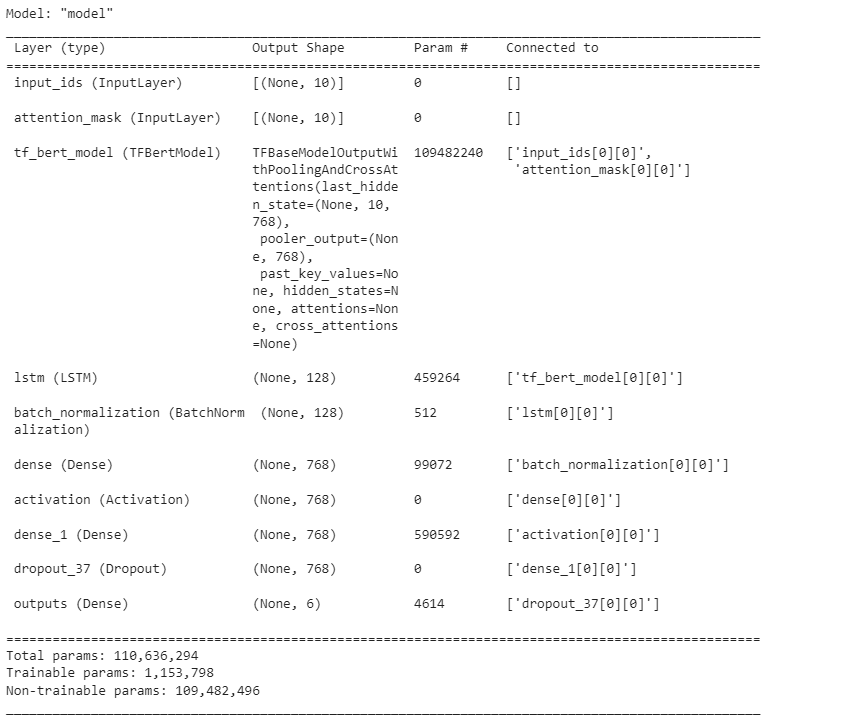}
    \caption{Proposed model summary}
    \label{fig:sum}
\end{figure}

We added a batch normalization layer to speed up training, and make learning easier, and a fully-connected output layer with a SoftMax function so that a probabilistic output of all tags for language identification would be produced. For further information, all the parameters and their summaries are depicted in Figure \ref{fig:sum}.
Figure \ref{fig:sum} shows our proposed model summary for word-level language identification in code-mixed Kannada-English texts.  

\subsection{Results}\label{res}
Table \ref{tab:modelres} depicts the overall results (official) of four experiments conducted in this work. From four experiments, using \textit{bert-base-uncased} in the embedding layer with the LSTM model out-performs other pre-trained languages models used in the embedding layer with the LSTM model with a weighted score of 0.85 precision, 0.84 recall, 0.84 F1-scores and a micro score of 0.62 precision, 0.62 recall, 0.61 F1-scores.

\begin{table}[h!]
\resizebox{\columnwidth}{!}{%
\begin{tabular}{lcccccc}
\hline
 & \multicolumn{3}{l}{\textbf{Weighted Score}} & \multicolumn{3}{l}{\textbf{Macro score}} \\
\multirow{-2}{*}{\textbf{Model}} & \textbf{P} & \textbf{R} & \textbf{F1-score} & \textbf{P} & \textbf{R} & \textbf{F1-score} \\ \hline
\multicolumn{1}{c}{bert-base-multilingual-uncased} & 0.83 & 0.82 & 0.82 & 0.62 & 0.57 & 0.57 \\
\multicolumn{1}{c}{xlm-roberta-large} & 0.84 & 0.85 & 0.84 & 0.64 & 0.59 & 0.61 \\
\multicolumn{1}{c}{roberta-base} & 0.83 & 0.8 & 0.81 & 0.63 & 0.55 & 0.52 \\
\multicolumn{1}{c}{\textbf{bert-base-uncased}} & {\textbf{0.85}} & {\textbf{0.84}} & {\textbf{0.84}} & {\textbf{0.62}} & {\textbf{0.62}} & {\textbf{0.61}} \\ \hline
\end{tabular}}
\caption{Performance of our models on the test set (official results)}
\label{tab:modelres}
\end{table}

The official rank of the top three teams participating in the shared task of word-level language identification in code-mixed Kannada-English texts is shown in Table \ref{tab:rank}. As shown in Table \ref{tab:rank} our model ranked second in overall results among all participant teams.

\begin{table}[h!]
\resizebox{\columnwidth}{!}{%
\begin{tabular}{clcccccc}
\hline
\multicolumn{1}{l}{} &  & \multicolumn{3}{l}{\textbf{Weighted Score}} & \multicolumn{3}{l}{\textbf{Macro score}} \\
\multicolumn{1}{l}{\multirow{-2}{*}{\textbf{Rank}}} & \multirow{-2}{*}{\textbf{Team name}} & \textbf{P} & \textbf{R} & \textbf{F1-score} & \textbf{P} & \textbf{R} & \textbf{F1-score} \\ \hline
 
1 & \multicolumn{1}{c}{tiya1012} & 0.87 & 0.85 & 0.86 & 0.67 & 0.61 & 0.62 \\
2 & \multicolumn{1}{c}{\textbf{Our team}} & \textbf{0.85} & \textbf{0.84} & \textbf{0.84} & \textbf{0.62} & \textbf{0.62} & \textbf{0.61} \\
 
2 & \multicolumn{1}{c}{Habesha} & 0.85 & 0.83 & 0.84 & 0.66 & 0.6 & 0.61 \\
 
3 & \multicolumn{1}{c}{lidoma} & 0.83 & 0.83 & 0.83 & 0.64 & 0.56 & 0.58\\ \hline
\end{tabular}}
\caption{Official rank of top 3 teams}
\label{tab:rank}
\end{table}
\begin{figure}[h!]
    \centering
    \includegraphics[width=\linewidth]{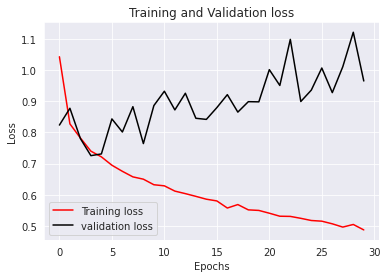}
    \caption{Training and validation loss of BERT based approach}
    \label{fig:loss}
\end{figure}

\begin{figure}[h!]
    \centering
    \includegraphics[width=\linewidth]{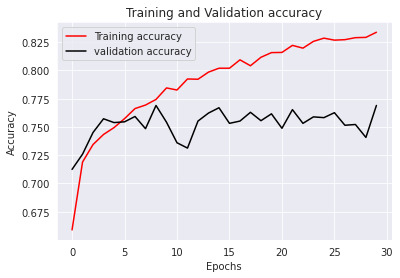}
    \caption{Training and validation accuracy of BERT based approach}
    \label{fig:acc}
\end{figure}

Figures \ref{fig:loss} and \ref{fig:acc} display the training and, validation losses, training, and validation accuracy of the BERT-based approach for code-mixed language identification tasks.  It is seen that the BERT-based model’s training loss decreases and stabilizes at a specific point, but the validation loss is not as stable as the training loss. This shows that the more specialized the model becomes with training data, the worse it is able to generalize to new data, resulting in an increase in generalization error.

The above result demonstrates that transformer-based models can give promising results when applied to NLP tasks like word-level language identification in code-mixed texts without considering any linguistic features.

\section{Conclusion}\label{con}
In this paper, we explored the application of BERT-based pre-trained language models to
identify languages at the word level in code-mixed data for Kannada-English texts. Pre-trained models with a combination of the LSTM model and a BERT-based model outperformed the others and have shown promising results in identifying languages in code-mixed Kannada-English texts. Our team achieved the second place in CoLI-Kanglish: word-level language identification in the code-mixed Kannada-English texts competition.
\section*{Acknowledgements}
The work was done with partial support from the Mexican Government through the grant A1S-47854 of CONACYT, Mexico, grants 20220852, 20220859, and 20221627 of the Secretaría de Investigación y Posgrado of the Instituto Politécnico Nacional, Mexico. The authors thank the CONACYT for the computing resources brought to them through the Plataforma de Aprendizaje Profundo para Tecnologías del Lenguaje of the Laboratorio de Supercómputo of the INAOE, Mexico and acknowledge the support of Microsoft through the Microsoft Latin America PhD Award.
\bibliography{anthology,custom}
\bibliographystyle{acl_natbib}




\end{document}